%% file: main_new.tex
\documentclass[letterpaper, 10 pt, conference]{ieeeconf}  

\IEEEoverridecommandlockouts                              
\input{./latexGoodPractices/preamble}

\title{\LARGE \bf 
    Exposing the Unseen: Exposure Time Emulation for Offline Benchmarking of Vision Algorithms
}

\author{Olivier Gamache$^{1}$, Jean-Michel Fortin, Mat\v ej Boxan, Maxime Vaidis, François Pomerleau, Philippe Giguère$^{1}$
\thanks{*This research was supported by Fonds de Recherche du Québec Nature et technologies (FRQNT) Team grant 254912 and Natural Sciences and Engineering Research Council of Canada (NSERC) DRC Grant through the grant CRD 538321-18, in collaboration with FP Innovations and Resolute Forest Products.}
	\thanks{$^{1}$Northern Robotics Laboratory, Université Laval, Québec City, Québec, Canada
		{\texttt{\small olivier.gamache@norlab.ulaval.ca}} and \texttt{\small {philippe.giguere@ift.ulaval.ca}}}%
}

\usepackage[switch]{lineno}
\usepackage[font=small]{caption}
\usepackage{subcaption}

\acrodef{SLAM}{Simultaneous Localization And Mapping}
\acrodef{VSLAM}{Visual Simultaneous Localization And Mapping}
\acrodef{VO}{Visual Odometry}
\acrodef{AE}{Automatic-Exposure}
\acrodef{CRF}{Camera Response Function}
\acrodef{DN}{Digital Number}
\acrodef{SNR}{Signal-to-Noise Ratio}
\acrodef{HDR}{High Dynamic Range}
\acrodef{IMU}{Inertial Measurement Unit}
\acrodef{FPS}{frames per second}
\acrodef{API}{Application Programming Interface}
\acrodef{GP}{Gaussian Process}
\acrodef{RMSE}{Root-Mean-Square Error}
\acrodef{ICP}{Iterative Closest Point}
\acrodef{SIFT}{Scale-Invariant Feature Transform}
\acrodef{RPE}{Relative Pose Error}
\acrodef{GNSS}{Global Navigation Satellite System}

\def\numberTrajectories{\num{55}}
\def\kilometerTravelled{10}
\def\totalHours{\num{5}}
\def\numberImages{\num{393238}}
\def\numberExposureBracketing{six}

\def\selectionMethod{\textsc{HigherNoSat}}
\def\datasetName{BorealHDR}

\def\BracketList{\Delta T_\text{bracket}}

\begin{document}
\maketitle
\thispagestyle{empty}
\pagestyle{empty}

\begin{abstract}

\ac{VO} is one of the fundamental tasks in computer vision for robotics.
However, its performance is deeply affected by \ac{HDR} scenes, omnipresent outdoor. 
While new \ac{AE} approaches to mitigate this have appeared, their comparison in a reproducible manner is problematic. 
This stems from the fact that the behavior of \ac{AE} depends on the environment, and it affects the image acquisition process.
Consequently, \ac{AE} has traditionally only been benchmarked in an online manner, making the experiments non-reproducible.
To solve this, we propose a new methodology based on an emulator that can generate images at any exposure time.
It leverages \datasetName,~a unique multi-exposure stereo dataset collected over \SI[detect-weight=true,mode=text]{\kilometerTravelled}{\kilo\meter}, on \numberTrajectories~trajectories with challenging illumination conditions.
Moreover, it includes lidar-inertial-based global maps with pose estimation for each image frame as well as \ac{GNSS} data, for comparison.
We show that using these images acquired at different exposure times, we can emulate realistic images, keeping a \ac{RMSE} below \SI[detect-weight=true,mode=text]{1.78}{\percent} compared to ground truth images. 
To demonstrate the practicality of our approach for offline benchmarking, we compared three state-of-the-art \ac{AE} algorithms on key elements of \ac{VSLAM} pipeline, against four baselines.
Consequently, reproducible evaluation of \ac{AE} is now possible, speeding up the development of future approaches.
Our code and dataset are available online at this link: \url{https://github.com/norlab-ulaval/BorealHDR}
\end{abstract}



\section{Introduction}
\label{sec:introduction}

\begin{figure}[tbp]
	\centering
	\includegraphics[width=0.48\textwidth]{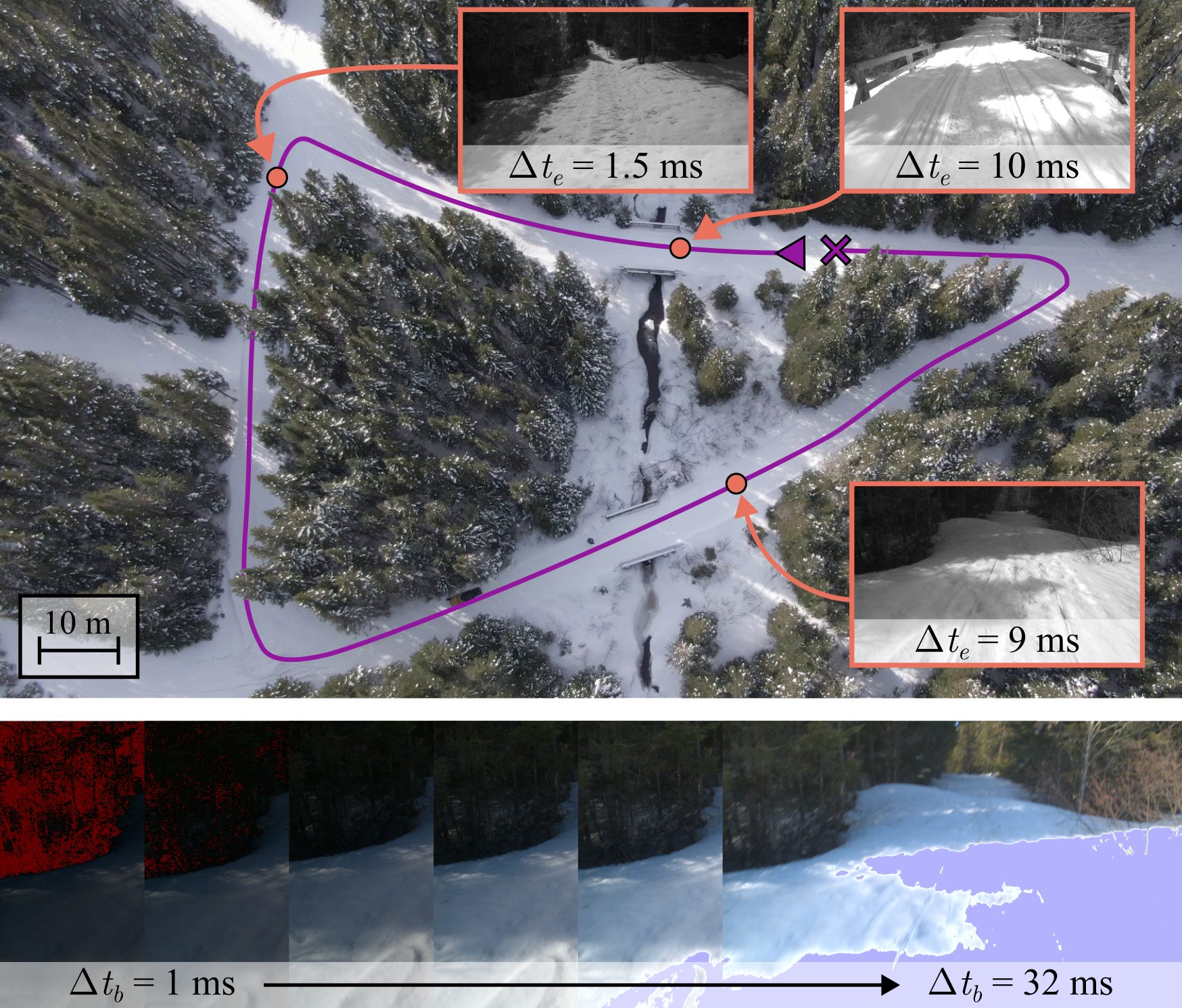}
	\caption{
            \textit{Upper image}: Overhead view of a trajectory from \datasetName~dataset, taken in Montmorency Forest, Québec, Canada.
            The traveled trajectory is shown in purple. 
            Possible emulated exposure times $\Delta t_{e}$ along the trajectory are depicted in orange.
            \textit{Lower image}: Acquired brackets (6), with one used to generate $\Delta t_e = \SI{9}{\milli\second}$ from the upper image.
            The bracket exposure times $\Delta t_b$ are $\{1, 2, 4, 8, 16, 32\}$ \si{\milli\second}, increasing the dynamic range of our capture by 30 dB or 5 stops. 
            Blue and red colorized pixels are over-exposed and under-exposed, respectively.
            }
	\label{fig:intro}
 \vspace{-0.1in}
\end{figure}

Cameras can capture high-resolution details of a scene, at high frame rates, and in a cost-effective manner.
Because of this, they are used in many robotic applications, ranging from object detection to localization.
One such application is \acf{VO}, which is the task of predicting the displacement of a camera between consecutive images.
It is the basis of many vision-based localization algorithms, such as \acf{VSLAM}, and improving such a task is still an active field of research~\citep{Han2023, Wang2022}. 
In outdoor settings, the quick variations in illumination level and \acf{HDR} scenes  can severely compromise the performances of \ac{VO} algorithms~\citep{Chahine2018}.
In one second, a car leaving a tunnel can experience an illumination variation over \SI{120}{\decibel}~\citep{Tomasi2021}, while a standard 12-bit channel camera has a theoretical dynamic range of around \SI{72}{\decibel}.
A boreal forest in winter is another example of such \ac{HDR} environments, where the sun reflected on the snow is highly contrasted with the darkness of the trees.
This contrast inevitably leads to saturated pixels on both ends of the spectrum, as highlighted by blue and red colorized hues in~\autoref{fig:intro}, resulting in the loss of valuable information for \ac{VO} algorithms.
Accordingly, researchers developed \acf{AE} approaches to modify the camera's parameters, such as exposure time and gain, during operation to reduce the impact of the scene's dynamic range on \ac{VO}~\citep{Zhang2017,Kim2020,begin2022}.

Unfortunately, given the online nature of exposure control, comparing each method is very challenging~\citep{Shin2019, Shim2019}, as one needs to exactly replicate test conditions for each run, including environment lighting and camera poses.
A frequent comparison method for \ac{AE} algorithms involves capturing images with as many cameras as there are methods to benchmark.
As the variety of tested methods grows, managing the experimental setup becomes increasingly complex and expensive. 
Another alternative involves replicating identical trajectories multiple times for each evaluated \ac{AE}, which becomes impractical for lengthy trajectories.
Moreover, in an uncontrolled environment such as the one illustrated in~\autoref{fig:intro}, the illumination constantly changes, making it impossible to have a replicable benchmark.

In this work, we propose a novel method for an equitable comparison between \ac{AE} algorithms applied to robot localization, based on the bracketing technique~\citep{gupta2013fibonacci} and realistic image emulation.
As illustrated by $\Delta t_e$ in~\autoref{fig:intro}, our emulator framework allows generating images at any desired exposure time for recorded trajectories. 
This is possible due to our multi-exposure dataset, called \datasetName, recorded in multi-seasonal conditions in Québec's forests and on Laval University's Campus.
It is composed of \numberTrajectories~trajectories, totaling more than \totalHours~\si{\hour} of recordings over \kilometerTravelled~\si{\kilo\meter}.
It also includes 3D lidar scans, inertial measurements, and \acf{GNSS} data.
The novelty of our dataset and our emulation technique allows us to benchmark \ac{AE} algorithms, feature extraction and \ac{VO} pipelines \textbf{in an offline manner}, as opposed to single-exposure datasets.
In short, our contributions are the following:
\begin{enumerate}
    \item A new comparison method for \ac{AE} algorithms based on an emulation framework that allows to generate realistic images at any plausible exposure time;
    \item A multimodal dataset that provides stereo images at multiple exposures, alongside 3D map references, in kilometer-scale \ac{HDR} environments; and
    \item First offline benchmarking of seven \ac{AE} algorithms on \ac{VO}-based experiments.
\end{enumerate}

\section{Related Work}
\label{sec:related_work}

We first review the existing methodologies for comparison of \ac{AE} algorithms.
Then, we go over available datasets for \ac{VO} algorithms and their limitations in the context of testing \ac{AE}, highlighting the advantages of our dataset.
Finally, we discuss important state-of-the-art \ac{AE} algorithms that will be used for the first benchmarking on a large \ac{VO} dataset.

\subsection{Methodologies for \ac{AE} Algorithms Comparison}
\label{sec:related_work_methodology}

The nature of \ac{AE} algorithms makes their comparison very challenging, as they actively change camera parameters during execution.
Here, we present state-of-the-art \ac{AE} benchmarking methods that address this issue, and categorize them between static and moving cameras approaches. 

\textbf{Static Cameras --} One way to evaluate \ac{AE} methods is to fix a camera and acquire multiple pictures at different exposure times.
\citet{Shim2014} used a surveillance camera to collect images with \num{210} combinations of exposure parameters, six times a day, to compare their new \ac{AE} method with others. 
\citet{Zhang2017} developed an \num{18} real-world static scenes dataset, where images at varying exposure times were collected in each scene.
\citet{Kim2018} collected a dataset made of \num{1000} images with multiple different exposures, covering indoor scenes and outdoor scenes.
\citet{Shin2019} were the only ones using a stereo camera to collect a static scene dataset.
Multi-exposed static scene images allow comparing some proxies of \ac{VO} such as single image feature detection. 
However, complete trajectories, as proposed in our multi-exposure dataset, are required to benchmark \ac{AE} algorithms on complete \ac{VO} pipelines.

\textbf{Moving Cameras --} Three trends can be observed using cameras in motion to compare \ac{AE} algorithms: simulation-based methods, multi-camera methods, and multi-trajectory methods.
An example of \emph{simulation-based methods} was used by ~\citet{Zhang2017}, which used the Multi-FoV synthetic dataset ~\citep{zhang2016benefit}, to simulate multiple versions of an image, but with different exposition levels.
Similarly, ~\citet{gomez2018learning} trained a neural network to produce images with higher gradient information using a synthetic dataset, where they simulated 12 different exposures.
Although these techniques reduce dataset development time, they face the \textit{Sim2Real} gap~\citep{hofer2021sim2real}, in opposition to our approach, which relies only on real-world images.
\emph{Multi-camera methods} are the most widespread for \ac{AE} algorithms comparison, leveraging a hardware solution for effective data collection.
In this setup, two or three cameras are typically installed on a mobile platform~\citep{Zhang2017,Shim2019,Kim2020,Mehta2020}.
This allows exact comparison since all \ac{AE} methods are executed simultaneously, thus facing the same conditions.
\citet{Wang2022} used four stereo cameras, facilitating \ac{VO} comparison, as the stereo depth estimation provides scaling information.
An important drawback for these multi-camera methods is that the hardware complexity grows with the number of \ac{AE} methods tested, rapidly becoming impractical.
On the contrary, our approach can be used to compare on an unlimited number of \ac{AE} methods, at a fixed acquisition cost. 
\emph{Multi-trajectory methods} consist of repeating multiple times the exact same trajectory using a different exposure control scheme at each iteration.
It was used by~\citet{begin2022} to develop a gradient \ac{AE} technique, which was tested using two cameras installed on a motorized \SI{0.16}{\meter\squared} motion table.
This allowed for exact repetitions of the same trajectories multiple times, with a ground truth precision near \SI{0.1}{\milli\meter}.
To evaluate \ac{AE} methods in the presence of motion blur, \citet{Han2023} acquired images from three cameras in motion on a motorized rail.
While providing precise ground truth, the rail approach severely restrains the total area covered by the trajectories.
It also limits the number of observable environments.
A variation of the multi-trajectory method was developed by~\citet{Kim2020}.
They drove using stop-and-go maneuvers on a single trajectory, where they stopped six times in total.
At each stop, they collected images at multiple exposure times.
This technique restricts the number of images that can be acquired, since any changes in the environment would corrupt the benchmark.

In our case, we do not make any static assumption, and we are able to compare \ac{AE} methods using standard \ac{VO} pipelines.
Our dataset has the advantage of being versatile and independent of the tested solutions, resulting in better replicability, even if new algorithms are developed.

\subsection{Datasets for \ac{VO}}
\label{sec:related_work_dataset}

Several datasets were collected aiming at improving \ac{VO} against challenging illumination conditions.
The KITTI~\citep{KITTI2012} and the Oxford~\citep{Oxford2020} datasets both acquired stereo cameras, lidars, \ac{IMU}, and \ac{GNSS} data, with Oxford offering different weathers, seasons, and illuminations.
The North Campus dataset~\citep{NorthCampus} also acquired urban stereo images, changing between indoors and outdoors, on a 15-month range, using a segway robotic platform.
The UMA-VI dataset~\citep{UMA-VI} had for main purpose to acquire \ac{HDR} images with a large number of low-textured scenes.
They used a handheld custom rig equipped with cameras and \ac{IMU}, but they only provided ground truth through loop closure error.
Closer to our dataset environment, TartanDrive~\citep{TartanDrive2022} and FinnForest~\citep{FinnForest2020} both acquired off-road data.
The TartanDrive dataset contains seven proprioceptive and exteroceptive modalities used for reinforcement learning purposes, including stereo images.
Although simpler, the FinnForest dataset is also composed of stereo images in summer and winter, showing the same forest landscapes under multiple conditions. 
From the papers described in~\autoref{sec:related_work_methodology}, only~\citep{Shin2019} and~\citep{begin2022} published their dataset. 
While the presented datasets allowed great improvements of \ac{VO} algorithms, ours is complementary by providing the full dynamic range of scenes through exposure times cycling. 
Combined with our emulation framework, we unlock the possibility to select the exposure time \emph{during playback}, expanding the realm of camera parameters algorithms evaluation.

\subsection{\ac{AE} Algorithms}
\label{sec:related_work_ae}

Vision-based localization necessitates sufficient contrast in images to maintain accuracy and robustness.
This, in turns, depends heavily on the camera's \ac{AE} algorithm and its accompanying target metric.
For instance,~\citet{Shim2014} used image gradient magnitude as such metric.
Their exposure control scheme generates seven synthetic versions of the latest acquired image, simulating different exposure levels, to identify the next exposure value maximizing their metric.
The \ac{AE} algorithm proposed by~\citet{Zhang2017} instead sorts the gradient level of each pixel, and applies a weight factor based on their percentile value.
By combining their quality metric and the \ac{CRF}, they predict the best next exposure value.
\citet{Kim2018} developed an image quality metric based on the gradient level and Shannon entropy~\citep{shannon1948mathematical}, used to detect saturation.
To demonstrate the benchmarking capabilities of our approach, we provide an implementation of the above methods, which are often used for comparison~\citep{begin2022,Wang2022}, since they cover main aspects of image quality for localization algorithms. 
Other techniques exist, such as~\citet{Shin2019}, who takes into account the \ac{SNR}, but we did not implement it, since it is mostly based on camera gain.

\section{Theory}
\label{sec:theory}

In this section, we detail the selected approaches for the development of our system, namely the emulation technique, the \ac{AE} implementations, and the lidar-based reference trajectories. 
Details on how our system can be used for benchmarking are presented in~\autoref{sec:results}.

\subsection{Emulation Technique}
\label{sec:theory_emulation}

From the real images acquired using the bracketing technique, we are able to emulate an image at any other realistic exposure time.
Our emulation method is based on the image acquisition process, which maps the scene radiance $E$ to the image pixel values $I(x)$, using the vignetting $V(x)$, the exposure time $\Delta t$, and the \ac{CRF} $f(\cdot)$.
This process can be expressed using the following equation~\citep{Bergmann2018}:
\begin{equation}
    I(x) = f\left(\Delta t V(x) E \right).
    \label{eq:photometric}
\end{equation}
From~\autoref{eq:photometric}, the relationship between two images $I_\text{source}$ and $I_\text{target}$ and their respective exposure times $\Delta t_\text{source}$ and $\Delta t_\text{target}$ can be defined as
\begin{equation}
    I_\text{target} = f\left(\frac{\Delta t_\text{target}}{\Delta t_\text{source}}\cdot f^{-1}\left(I_\text{source}\right)\right),
    \label{eq:emulation}
\end{equation}
where $f^{-1}(\cdot)$ is the inverse \ac{CRF}~\citep{Grossberg2003}.
To estimate $f(\cdot)$ and $f^{-1}(\cdot)$, we take multiple images of a static scene at several exposure times, allowing to capture the whole dynamic range at fixed radiance, following~\citep{engel2016monodataset}.
With $f^{-1}(\cdot)$ and~\autoref{eq:emulation}, it is thus possible to emulate any targeted exposure times $I_\text{target}$ by using a known image with an exposure time $\Delta t_\text{source}$.
Note that, for the sake of simplicity, an image taken from one exposure time in the bracketing cycle will be called \textit{bracket}.

Considering that the data is acquired while moving, the brackets are not taken at the same time nor position. 
Therefore, we decided not to interpolate between the brackets, since it would create artifacts in the resulting image.
Instead, we select the most appropriate $I_\text{source}$ from the available brackets to emulate $I_\text{target}$.
The selection process takes into account the distance between $\Delta t_\text{source}$ and $\Delta t_\text{target}$, and the amount of saturated pixels in $I_\text{source}$, since these do not contain any usable information.
Also, if we select a bracket such as $\Delta t_\text{target}/\Delta t_\text{source} > 1$, the result will be an image with higher pixel values, which means that the noise in the signal will also increase.
Therefore, it is preferable to select $I_\text{source}$ that has a higher exposure time compared to our desired $I_\text{target}$ to maximize the \ac{SNR}.

Based on these considerations, we developed a simple selection method named \selectionMethod.
The first step is to find the two closest brackets $\Delta t_\text{bl}$ and $\Delta t_\text{bh}$, that bound $\Delta t_\text{target}$, such that: 
\begin{equation}
    \Delta t_\text{bl} < \Delta t_\text{target} < \Delta t_\text{bh}.
    \label{eq:bound}
\end{equation}
Then, for each of the two brackets, we calculate the amount of image saturation, which corresponds to the number of pixels with values of \num{0} and \num{4095}, for our \num{12}-bit channel images.
If the saturation level of the higher bracket is below \SI{1}{\percent}, we select it as the best candidate, otherwise, we pick the lower one.
Finally, if $\Delta t_\text{target}$ is outside the range of available $\Delta t_\text{source}$, we select the closest bracket.

\subsection{Implementation Details of Compared \ac{AE} Algorithms}
\label{sec:theory_implementation}

Based on~\autoref{sec:related_work_ae}, we implement and benchmark three state-of-the-art \ac{AE} methods: $M_\text{Shim}$~\citep{Shim2014}, $M_\text{Zhang}$~\citep{Zhang2017}, and $M_\text{Kim}$~\citep{Kim2018}.
These methods are not open-source, thus they were all implemented based on our understanding of the papers.
$M_\text{Zhang}$ suffers from instabilities coming from the \ac{CRF} estimation, which were partly resolved using the description in~\citep{Shin2019}.
The exposure control scheme of $M_\text{Kim}$ used a \ac{GP}, which sparsely sweep the camera exposure parameters until convergence.
In our implementation, we use a sliding window on the trajectory to only consider the most recent images in the \ac{GP} training.
This can cause steep changes in the desired exposure time, since the exploration algorithm of the \ac{GP} needs to cover a wide range of values to converge.
We also implemented four baseline methods, corresponding to the typical \ac{AE} algorithms.
The first baseline algorithm is a fixed exposure time approach $M_{fixed}$.
The exposure time is selected once, at the beginning of each sequence, by using a brightness target of \SI{50}{\percent}.
The three other baselines are variable exposure algorithms, seeking to keep the average brightness target to \SI{30}{\percent}, \SI{50}{\percent}, and \SI{70}{\percent} of the \num{12}-bit depth range. 
They are named respectively $M_{\SI{30}{\percent}}$, $M_{\SI{50}{\percent}}$, $M_{\SI{70}{\percent}}$.
These percentages were chosen as to cover a wide range of exposure.

\subsection{Reference trajectories}
\label{sec:theory_ground_truth}

In addition to the cameras, the platform is equipped with a lidar and a \ac{GNSS} antenna, which both provide useful information to estimate the traveled trajectories.
In boreal forests,~\citet{Baril2022} demonstrated that \ac{GNSS} signals are unreliable, due to attenuation caused by the dense vegetation canopy.
Therefore, our reference trajectories were generated by employing a low-drifting lidar-inertial-\ac{SLAM} variant~\citep{Kubelka2022}, based on the libpointmatcher library\footnote{\url{https://github.com/norlab-ulaval/libpointmatcher}}.
This method registers deskewed point clouds, at a rate of \SI{10}{\hertz}, into a dense 3D map of the environment.
It relies on the on-board lidar and is illumination-independent, with median error values around \SI{1}{\percent} of the total trajectory length.
Note that, we do not assume that the reference trajectories are equivalent to the ground truths, but that they provide an adequate precision for comparison with our \ac{VO} pipeline, as explained in~\autoref{sec:results_benchmark_vo}.

\section{Results}
\label{sec:results}

This section first describes our acquisition platform, before evaluating the performances of the emulation method.
Then, we elaborate our data collection pipeline, and finally, conduct the first online \ac{AE} benchmarking on large \ac{VO}-based trajectories.

\subsection{Experimental Setup}
\label{sec:results_experimental_setup}

\begin{figure}
    \centering
    \vspace{0.1in}
    \includegraphics[width=0.47\textwidth]{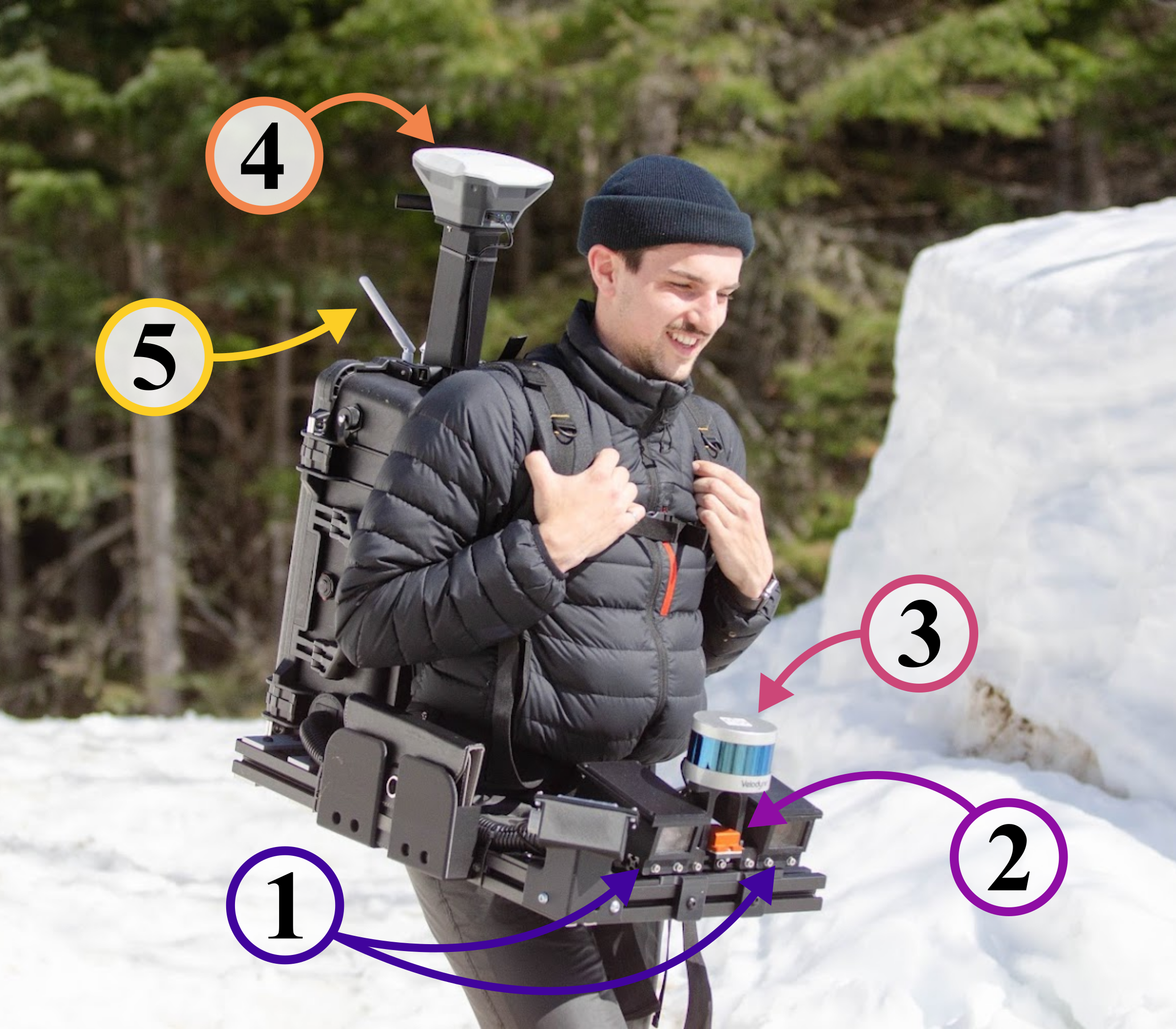}
    \caption{Picture of the developed backpack for the dataset acquisition. Main components are identified as follows: \raisebox{.5pt}{\textcircled{\raisebox{-.9pt} {1}}} Two Basler a2A1920-51gcPRO cameras, \raisebox{.5pt}{\textcircled{\raisebox{-.9pt} {2}}} Xsens MTI-30 \ac{IMU}, \raisebox{.5pt}{\textcircled{\raisebox{-.9pt} {3}}} VLP16 3D lidar, \raisebox{.5pt}{\textcircled{\raisebox{-.9pt} {4}}} Emlid Reach RS+ \ac{GNSS} receiver, and \raisebox{.5pt}{\textcircled{\raisebox{-.9pt} {5}}} Ubiquiti UniFi UAP-AC-M wifi antenna.}
    \label{fig:backpack}
\end{figure}

The \datasetName~dataset was collected using a homemade water-resistant acquisition backpack, depicted in~\autoref{fig:backpack}.
The enclosure is a Pelican case \num{1510} containing a battery and a Jetson AGX Xavier Development kit embedded computer. 
The camera sensors are two Basler a2A1920-51gcPRO, in a stereo-calibrated configuration with a baseline of \SI{18}{\centi\meter}, and hardware-triggered by an STM32 microcontroller ensuring precise exposure times and sensor synchronization.
In addition, the platform is equipped with a Velodyne VLP-16 lidar, an Xsens MTI-30 \ac{IMU}, and an Emlid Reach RS+ \ac{GNSS} receiver.
Similar to~\citep{Jiajun2020}, the extrinsic calibrations between the sensors are given by the 3D model of the platform.
The intrinsic calibrations for the cameras were also calculated, by using a standard checkerboard.
The calibration sequences, and the 3D model are provided with the dataset.
The images are acquired at a rate $r_\text{real}$ of \num{22}~\ac{FPS} in \num{12}-bit color, using loss-less compression.
The image acquisition process cycles through \numberExposureBracketing~exposure time values, \ie{} $\BracketList=\{1,2,4,8,16,32\}$~\si{\milli\second}, yielding an effective emulation rate of $r_\text{emul}=$ 3.66 \ac{FPS}.
This number of brackets is a compromise between the offline $r_\text{emul}$ and the emulation error, detailed in~\autoref{sec:results_emulation}.
The $\BracketList$ values were chosen to cover the whole dynamic range of our target environments, while mitigating motion blur.
The exposure time is doubled at every step of the cycle, which is equivalent to a one stop increase in photography.
The small footprint of the acquisition platform allows collecting data in narrow and hard-to-access spaces for robotic vehicles, which was fundamental for our \kilometerTravelled~\si{\kilo\meter} dataset in \ac{HDR} environments.

\subsection{Emulation}
\label{sec:results_emulation}

\begin{figure*}[htbp]
	\centering
        \vspace{0.1in}
	\includegraphics[width=0.98\textwidth]{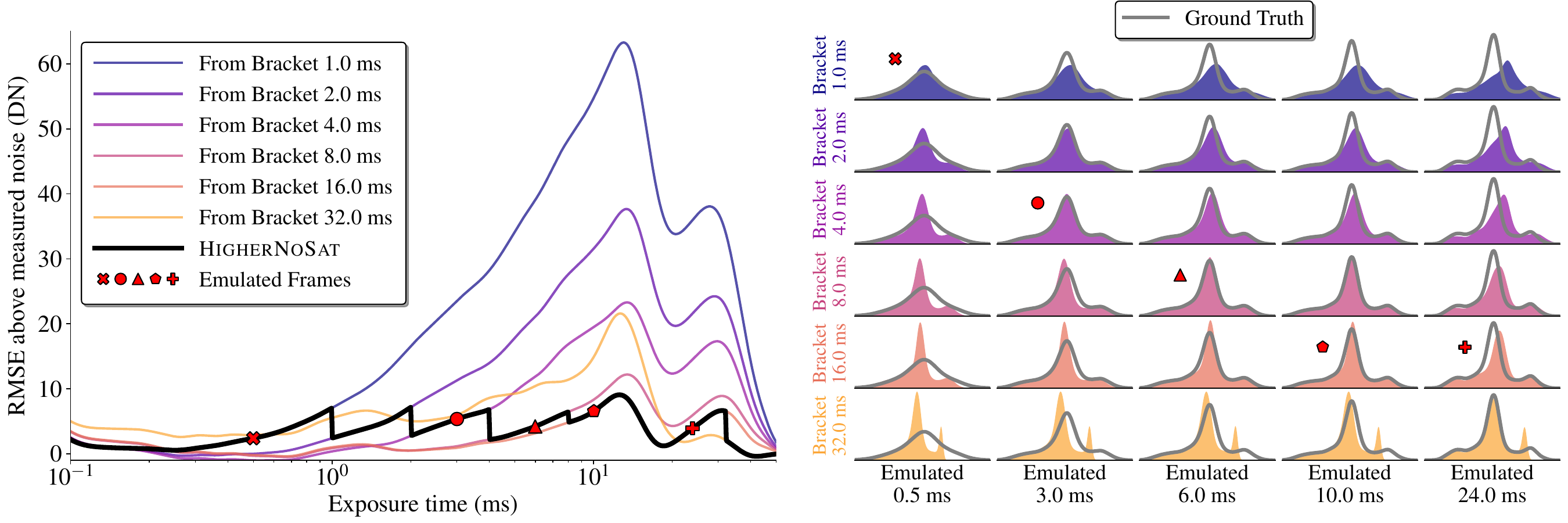}
	\caption{Validation of our emulation framework using \num{1000} ground truth images captured with constant illumination, but varying exposure time. \textit{\textbf{Left:}} \ac{RMSE} curves showing the emulation error if a single bracket was always selected, in colors, compared to our bracket selection method \selectionMethod, in black. Red symbols correspond to the emulated exposure times displayed on the right. \textit{\textbf{Right:}} Qualitative comparison of the distributions of five emulated exposure times (columns) from six different bracket images (rows). The markers are placed next to the selected bracket by \selectionMethod. Ground truth distributions are overlaid in gray for each column.
            }
	\label{fig:emulation_performances}
\end{figure*}
To evaluate the performances of our image emulator, we collect five test sequences of \num{1000} ground truth images with exposure times ranging from \SI{20}{\micro\second} to \SI{50}{\milli\second}, in static scenes both indoor and outdoor.
Then, we emulate the same images following the method described in~\autoref{sec:theory_emulation}, and calculate the \acf{RMSE} between $I_\text{emul}$ and $I_\text{GT}$ for each ground truth exposure.
To account for the camera's intrinsic noise, we average the error between \num{25} consecutive images at the same exposure time, over the whole spectrum, and subtract this noise from our results.
The five test sequences serve to validate our bracket selection method \selectionMethod, which does not have access to ground truth images in real world settings.
Overall, our tests show that the emulation method maintains a median \ac{RMSE} of \SI{0.21}{\percent} and never surpasses \SI{1.78}{\percent}, emphasizing its consistency and accuracy.
Note that the sequences are provided with the dataset.

One of the test sequences is detailed in~\autoref{fig:emulation_performances}, showing emulation performances in controlled lab settings.
On the left side, the \ac{RMSE} curve from \selectionMethod, in black, is compared to the error obtained by selecting a single bracket as $I_\text{source}$ for the whole range, in colors.
It shows that our bracket selection maintains the error close to the lowest value of the \numberExposureBracketing~colored curves.
The right side plots present a qualitative evaluation of the emulated images from the acquired brackets.
Each column exposes the distributions of pixel intensity for one target exposure time, while each row emulates this image from a different bracket.
Red markers represent which exposure time in $\BracketList$ was chosen by our emulator via \selectionMethod, for this image.
We observe that our bracket selector consistently picks the closest or second-closest emulated image to the ground truth distribution, overlaid in gray.

\subsection{Dataset Gathering}
\label{sec:results_dataset_gathering}

Our \datasetName~dataset comprises \numberTrajectories~sequences totaling around \kilometerTravelled~\si{\kilo\meter} for \totalHours~\si{\hour}, and around \SI{2.5}{\tera\byte} of data.
Half of the trajectories are loops, which is a common practice to estimate drift in \ac{VSLAM} datasets.
The relatively low $r_\text{emul}$ implies that the data should be collected at a slow walking speed to minimize displacements between each acquisition cycle. 
Hence, our average pace is around \SI{2}{\kilo\meter/\hour}, which increases the data collection time but does not impact the users of our offline emulation approach.
In total, \numberImages~images were gathered on four separate days in summer and winter \num{2023}.
The winter data, from April 20$^{\text{th}}$ to April 21$^{\text{st}}$, was acquired in the Montmorency boreal Forest near Québec City, Canada.
The summer trajectories were collected on the campus of Laval University and on Mont-Bélair, in Québec City, respectively on the 25$^{\text{th}}$ and the 27$^{\text{th}}$ of September.
There are 44 trajectories from Montmorency boreal Forest, four acquire on the university campus, and seven from Mont-Bélair, each site respectively totaled \SI{7.5}{\kilo\meter}, \SI{1}{\kilo\meter}, and \SI{1.5}{\kilo\meter}.
We chiefly focused the acquisitions for \ac{HDR} scenes, either by capturing trees and snow in the camera frame for April's data, or shadows and sunlight's area, in the September's ones.
In addition to \ac{HDR} scenarios, multiple trajectories contain challenging conditions for \ac{VO}, such as sun glares and texture-less environment from immaculate snow and a lake steady water.
In winter, we used snowshoes to get to remote locations, to collect not only different scene illuminations, but also various 3D structures, allowing for a richer collection of test scenarios. 


\subsection{Benchmark}
\label{sec:results_benchmark}

The main contribution of our paper is to unlock \emph{offline} testing of active vision methods that could only previously be tested in an \emph{online} manner. 
To this effect, we conduct two series of tests to benchmark the methods explained in~\autoref{sec:theory_implementation}, namely \emph{(1) feature tracking and uniformity}, and \emph{(2) stereo \ac{VO}}, both highlighting the impact of each \ac{AE} algorithm in key aspects of a \ac{VSLAM} pipeline.
\subsubsection{Feature Tracking and Uniformity}
\label{sec:results_benchmark_features}
\begin{figure}[htbp]
    \vspace{3mm}
    \centering
    \includegraphics[width=0.47\textwidth]{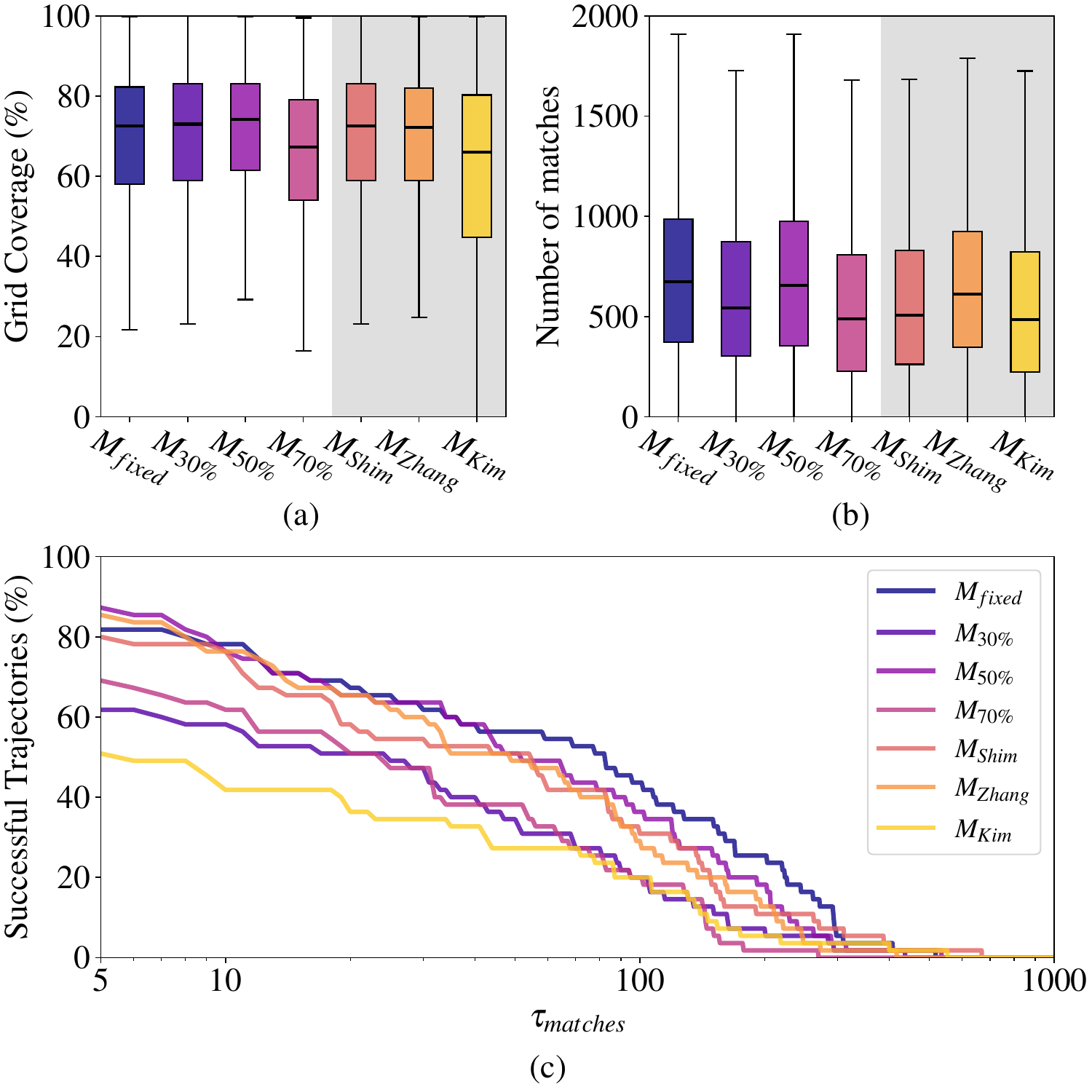}
    \caption{SIFT features analysis. (a) Uniformity of detected keypoints for an image divided into a $20\times20$ grid. (b) Number of match features between a pair of images for all \ac{AE} metrics.
    The gray shading in (a) and (b) is to highlight the state-of-the-art \ac{AE} methods.
    (c) Percentage of successful trajectories based on the number of matches detected. If one image contains less matches than $\tau_\text{matches}$, the sequence is marked as not successful.}
    \label{fig:keypoints}
    \vspace{-0.2in}
\end{figure}
For feature-based \ac{VO} algorithms, the performance is dependent on the quality and quantity of detected keypoints.
Therefore, we evaluate feature detection for each implemented \ac{AE} algorithm on our recorded dataset, over three distinct tests.
We selected the \ac{SIFT}~\citep{lowe1999object} feature-detector, since it is pervasive in the literature, and the most accurate according to~\citep{tareen2018comparative}.
Learning-based feature-detectors such as SuperPoint~\citep{detone2018superpoint} could be considered for evaluation.
However, their performance is sensitive to the domain gap between training and test data, which could bias results.
Consequently, \ac{SIFT} provides for an equitable analysis of the methods.
The first test consists of estimating the uniformity of the detected keypoints, by dividing each image into a $20\times20$ grid and assessing its occupancy rate.
A uniform distribution of features is presumed to be a good proxy for \ac{VO} algorithm performance.
The results are displayed in~\autoref{fig:keypoints}\textcolor{red}{a}, which shows that most implemented \ac{AE} methods yield a similar grid coverage.
We observe that $M_{\SI{70}{\percent}}$ and $M_{\text{Kim}}$ provide slightly reduced uniformity, with median values respectively \SI{67.3}{\percent} and \SI{66.0}{\percent} under the average of the other combined methods.
In~\autoref{fig:keypoints}\textcolor{red}{b} we show that we can also obtain the number of detected features that are matched between consecutive images, again for all implemented \ac{AE} methods.
$M_\text{fixed}$ is the one tracking the highest number of features between two images, with a median of \num{674} matches.
This result was expected since the exposure time is calibrated when starting each sequence, and the lack of exposure change keeps a better flow between consecutive images.
Now, looking at full trajectories, we define a success criterion corresponding to a minimal number of matches $\tau_\text{matches}$ between each consecutive image pairs. 
If, for a given \ac{AE} algorithm and trajectory, this criterion is always true, then the sequence is considered successful.
Accordingly, we evaluate the percentage of successful trajectories with $min(\tau_\text{matches}) = 5$, the minimum number of matches for motion estimation~\citep{nister2004efficient}, and the results are shown in~\autoref{fig:keypoints}\textcolor{red}{c}.
We observe that the overall robustness of the \ac{AE} algorithms decreases rapidly.
For $\tau_\text{matches}=100$, which is largely under any first quartile from~\autoref{fig:keypoints}\textcolor{red}{b}, $M_{\SI{30}{\percent}}$, $M_{\SI{70}{\percent}}$, and $M_{\text{Kim}}$ shows the poorest performance, completing \SI{20}{\percent} of the dataset sequences successfully, while the best method, $M_\text{fixed}$, achieves \SI{44}{\percent} of success.
This clearly shows the challenge that represents our dataset for \ac{VO}. 

Many of the developed benchmarking techniques for \ac{AE} algorithms would have been able to conduct our first two experiments.
However, the third one could only have been done using methodologies based on moving camera, since we evaluate the features on trajectories.
More importantly, our approach enables entirely reproducible testing, offline.
Consequently, we can always add more \ac{AE} methods to the benchmarks, without having to collect new data.

\subsubsection{Stereo \acf{VO}}
\label{sec:results_benchmark_vo}

\begin{figure}[htbp]
	\centering
	\includegraphics[width=0.48\textwidth]{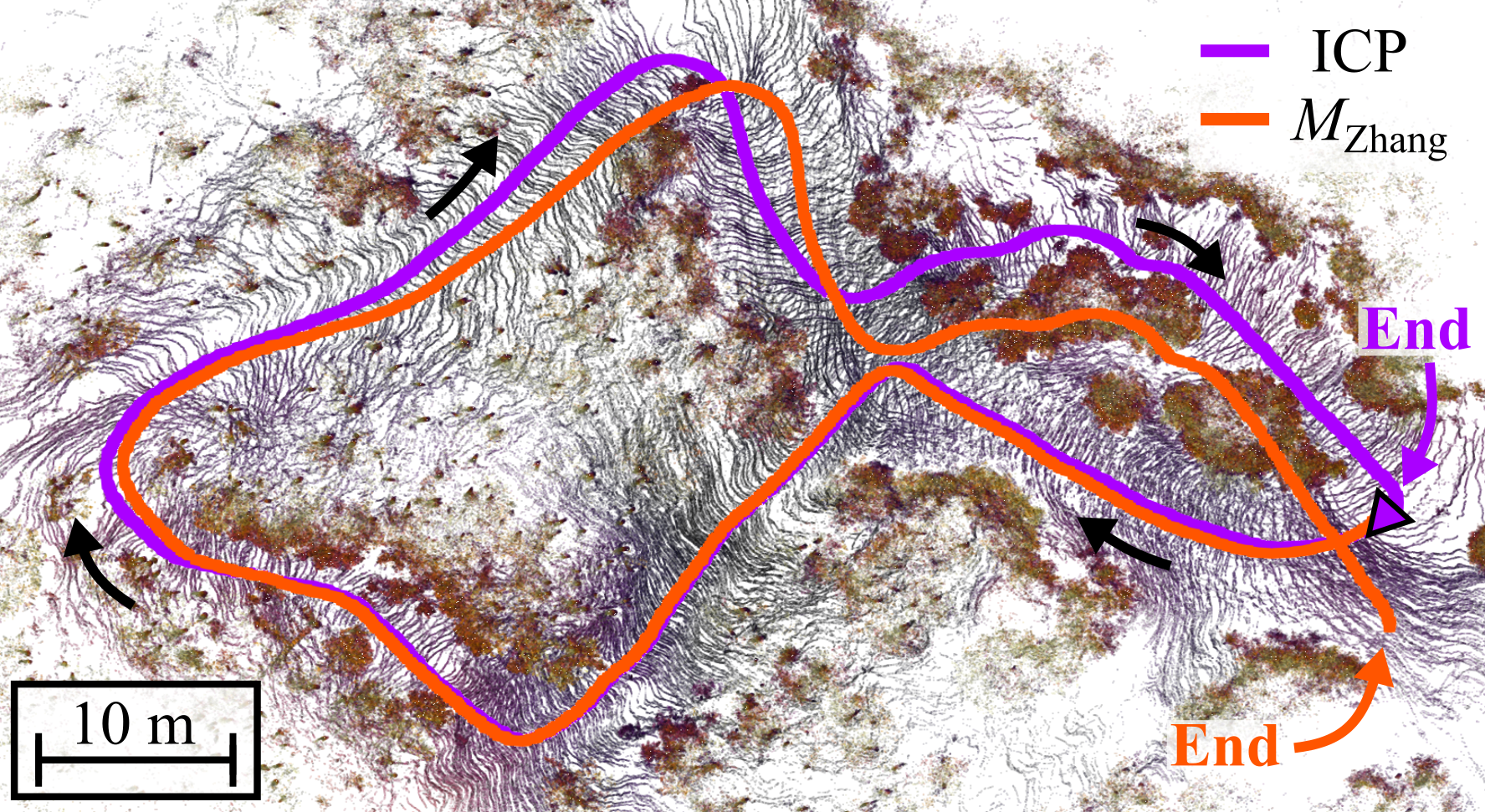}
	\caption{
            Qualitative example illustrating the trajectory \texttt{backpack\_2023-04-21-10-57-59} from our \datasetName~dataset.
            The lidar-based trajectory (purple) and $M_\text{Zhang}$'s emulated \ac{VO} result (orange) are illustrated superimposed on the generated lidar 3D map. The higher drift of the \ac{VO} localization can be highlighted with the loop closure done in this forest environment.
            Black color represents the ground, and other colors highlight structures in the environment.
            }
	\label{fig:results_lidar_gt}
\end{figure}

We also investigate the impact of \ac{AE} methods on \ac{VO}, which is a key component of \ac{VSLAM}.
We implemented a minimal stereo \ac{VO} pipeline based on \emph{OpenCV}~\citep{opencv_library} using SIFT \citep{lowe1999object} for feature detection.
The \emph{evo}\footnote{\url{https://github.com/MichaelGrupp/evo}} library is used to calculate the \ac{RPE}~\cite{sturm2012benchmark} for each \ac{AE} algorithm.
Our reference trajectories are generated using the lidar-inertial-\ac{SLAM} system, as explained in~\autoref{sec:theory_ground_truth}.
For an equitable comparison, the scale alignment feature from the \emph{evo} library was applied on the \ac{VO} trajectories.
An example using one trajectory from our dataset is illustrated in~\autoref{fig:results_lidar_gt}, along with the 3D map.
The \ac{VO} algorithm was unsuccessful for \num{11} of the \numberTrajectories~trajectories, resulting in \SI{20}{\percent} of failure cases, highlighting the challenge that represents our new dataset \datasetName~for \ac{VO}. 
Multiple of the failed trajectories contain sun glares and illumination variation, which could explain the poor performances.

\begin{figure*}[htbp]
	\centering
        \vspace{0.1in}
	\includegraphics[width=0.98\textwidth]{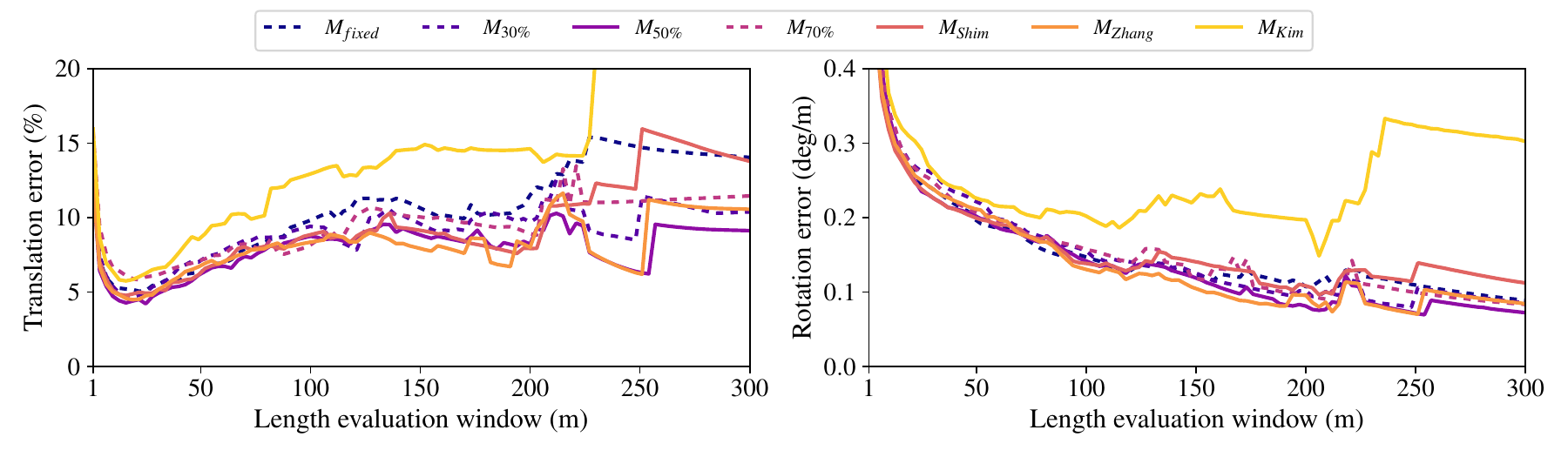}
	\caption{\ac{RPE} over all the trajectories as a function of the trajectory length, for the seven implemented \ac{AE} algorithms.
            \textit{Left}: Median translation error. \textit{Right}: Median rotation error.
            The line styles depict three clusters, based on the translation performances, going from the top with solid, dashed, and then solid again.
            }
	\label{fig:results_vo}
\end{figure*}
The \ac{RPE} is calculated on all the trajectories for which the stereo \ac{VO} pipeline converged, and is displayed in~\autoref{fig:results_vo}.
We observe that all the \ac{AE} methods have a minimal translation error above \SI{4.21}{\percent}, obtained by $M_{\SI{50}{\percent}}$.
The least performing method was $M_\text{Kim}$, which has a minimal error of \SI{5.75}{\percent}.
This is caused by the \ac{GP}'s exploration step, as explained in~\autoref{sec:theory_implementation}.
The rotational error has similar results, where $M_{\text{Zhang}}$ and $M_{\SI{50}{\percent}}$ are the two best methods of the benchmark, with a minimal error of \SI{0.07}{deg/\meter}.
Even though $M_\text{fixed}$ was among the best methods for feature tracking, its unresponsiveness to illumination changes results in an error of \SI{5.1}{\percent} in translation and \SI{0.09}{deg/\meter} in rotation, when applied to \ac{VO}.
From this experiment, all the tested \ac{AE} algorithms performed similarly on the dataset, with a small advantage for gradient-based methods.
In summary, our new developed approach allows for the first comparison of \ac{AE} methods directly on a large \ac{VO} dataset, in an offline manner.


\section{Conclusion And Future Work}
\label{sec:conclusion}

In this work, we proposed an emulator framework based on a multi-exposure dataset, which allows comparing \ac{AE} algorithms in an offline and reproducible manner.
Our dataset, \datasetName, contains \numberTrajectories~trajectories combining for \SI{\kilometerTravelled}{\kilo\meter}, focusing on collecting images in \ac{HDR} scenes taken in a snowy boreal forest environment.
In addition to the \num{12}-bit color stereo images, we also provide lidar-inertial-\ac{SLAM} based pose for each image and 3D map for each sequence, based on lidar data.
We have shown the versatility of our methodology by benchmarking seven \ac{AE} algorithms, of which three were from recent works.
We concluded that our emulation approach is an efficient solution for offline benchmarking of active algorithms, such as \ac{AE}, by making experiments on multiple key elements affecting \ac{VSLAM} pipelines.
In the future, we plan to extend our dataset to contain more drastic illumination changes and scenarios.
We also plan to design novel \ac{AE} approaches that take into account the pose uncertainty estimates obtained by the \ac{VSLAM} pipeline, which were previously difficult to develop without a flexible testing methodology.


\printbibliography

\end{document}

%% file: latexGoodPractices/preamble.tex
\usepackage[l2tabu,orthodox]{nag}

\usepackage[
    backend=bibtex8,
    style=ieee,
    sorting=none,
    natbib=true,
    doi=false,
    isbn=false,
    url=false,
    eprint=false,
    maxcitenames=1,
    mincitenames=1
]{biblatex}

\usepackage[pdftex,colorlinks]{hyperref}

\usepackage[printonlyused]{acronym}

\usepackage{siunitx}
\usepackage[all]{nowidow}

\usepackage[dvipsnames]{xcolor}

\usepackage{lipsum}

\usepackage[normalem]{ulem}

\usepackage{xspace} 
\newcommand{\ie}{i.e.,\xspace{}}

\usepackage[pdftex]{graphicx}

\usepackage{epstopdf}

\usepackage{import}

\graphicspath{{./latexGoodPractices/}}

\usepackage{booktabs}

\usepackage{tabularx}
\usepackage{multirow, multicol}

\usepackage{amssymb,amsfonts,amsmath,amscd}

\usepackage{bm}

\newcommand{\bbm}{\begin{bmatrix}}
\newcommand{\ebm}{\end{bmatrix}}

%% file: references.bib
@inproceedings{Chahine2018,
  year = {2018},
  author = {Georges Chahine and Cedric Pradalier},
  title = {{Survey of Monocular {SLAM} Algorithms in Natural Environments}},
  booktitle = {Conference on Computer and Robot Vision ({CRV})}
}

@inproceedings{gupta2013fibonacci,
  doi = {10.1109/iccv.2013.186},
  url = {https://doi.org/10.1109/iccv.2013.186},
  year = {2013},
  author = {Mohit Gupta and Daisuke Iso and Shree K. Nayar},
  title = {{Fibonacci Exposure Bracketing for High Dynamic Range Imaging}},
  booktitle = {{IEEE} International Conference on Computer Vision (ICCV)}
}

@article{hofer2021sim2real,
  title={{Sim2Real in robotics and automation: Applications and challenges}},
  author={H{\"o}fer, Sebastian and Bekris, Kostas and Handa, Ankur and Gamboa, Juan Camilo and Mozifian, Melissa and Golemo, Florian and Atkeson, Chris and Fox, Dieter and Goldberg, Ken and Leonard, John and others},
  journal={IEEE transactions on automation science and engineering},
  volume={18},
  number={2},
  pages={398--400},
  year={2021},
  publisher={IEEE}
}

@inproceedings{Shim2014,
   author = {Inwook Shim and Joon-Young Lee and In So Kweon},
   booktitle = {IEEE/RSJ International Conference on Intelligent Robots and Systems (IROS)},
   pages = {1011-1017},
   title = {{Auto-adjusting camera exposure for outdoor robotics using gradient information}},
   year = {2014},
}

@article{Shim2019,
   author = {Inwook Shim and Tae Hyun Oh and Joon Young Lee and Jinwook Choi and Dong Geol Choi and In So Kweon},
   doi = {10.1109/TCSVT.2018.2846292},
   issn = {10518215},
   issue = {6},
   journal = {IEEE Transactions on Circuits and Systems for Video Technology},
   month = {6},
   pages = {1569-1583},
   publisher = {Institute of Electrical and Electronics Engineers Inc.},
   title = {{Gradient-based camera exposure control for outdoor mobile platforms}},
   volume = {29},
   year = {2019},
}

@inproceedings{Mehta2020,
   author = {Ishaan Mehta and Mingliang Tang and Timothy D. Barfoot},
   booktitle = {Conference on Computer and Robot Vision (CRV)},
   pages = {166-173},
   title = {{Gradient-Based Auto-Exposure Control Applied to a Self-Driving Car}},
   year = {2020},
}

@inproceedings{Shin2019,
   author = {Ukcheol Shin and Jinsun Park and Gyumin Shim and Francois Rameau and In So Kweon},
   booktitle = {IEEE/RSJ International Conference on Intelligent Robots and Systems (IROS)},
   pages = {1165-1172},
   title = {{Camera Exposure Control for Robust Robot Vision with Noise-Aware Image Quality Assessment}},
   year = {2019},
}

@inproceedings{Zhang2017,
   author={Zhang, Zichao and Forster, Christian and Scaramuzza, Davide},
   booktitle = {IEEE International Conference on Robotics and Automation (ICRA)},
   pages = {3894-3901},
   title = {{Active exposure control for robust visual odometry in HDR environments}},
   year = {2017},
}

@article{Kim2020,
   author = {Joowan Kim and Younggun Cho and Ayoung Kim},
   doi = {10.1109/TRO.2020.2985597},
   issn = {1552-3098},
   journal = {IEEE Transactions on Robotics (T-RO)},
   pages = {1256-1271},
   title = {{Proactive Camera Attribute Control Using Bayesian Optimization for Illumination-Resilient Visual Navigation}},
   volume = {36},
   url = {https://ieeexplore.ieee.org/document/9098963/},
   year = {2020},
}

@inproceedings{Kim2018,
   author = {Joowan Kim and Younggun Cho and Ayoung Kim},
   booktitle = {IEEE International Conference on Robotics and Automation (ICRA)},
   pages = {857-864},
   title = {{Exposure Control Using Bayesian Optimization Based on Entropy Weighted Image Gradient}},
   year = {2018},
}

@inproceedings{gomez2018learning,
  title={{Learning-based image enhancement for visual odometry in challenging HDR environments}},
  author={Gomez-Ojeda, Ruben and Zhang, Zichao and Gonzalez-Jimenez, Javier and Scaramuzza, Davide},
  booktitle={IEEE International Conference on Robotics and Automation (ICRA)},
  pages={805--811},
  year={2018},
}

@article{Grossberg2003,
   author = {M.D. Grossberg and S.K. Nayar},
   doi = {10.1109/TPAMI.2003.1240119},
   issn = {0162-8828},
   issue = {11},
   journal = {IEEE Transactions on Pattern Analysis and Machine Intelligence},
   month = {11},
   pages = {1455-1467},
   title = {{Determining the camera response from images: What is knowable?}},
   volume = {25},
   url = {http://ieeexplore.ieee.org/document/1240119/},
   year = {2003},
}

@article{Tomasi2021,
   author = {Justin Tomasi and Brandon Wagstaff and Steven L. Waslander and Jonathan Kelly},
   doi = {10.1109/LRA.2021.3058909},
   issn = {2377-3766},
   journal = {IEEE Robotics and Automation Letters (RA-L)},
   pages = {2028-2035},
   publisher = {IEEE},
   title = {{Learned Camera Gain and Exposure Control for Improved Visual Feature Detection and Matching}},
   number={2},
   volume = {6},
   url = {https://ieeexplore.ieee.org/document/9353970/},
   year = {2021},
}

@article{begin2022,
   author={B{\'e}gin, Marc-Andr{\'e} and Hunter, Ian},
   doi = {10.3390/s22030835},
   issn = {1424-8220},
   issue = {3},
   journal = {Sensors},
   pages = {835},
   publisher = {MDPI},
   title = {{Auto-Exposure Algorithm for Enhanced Mobile Robot Localization in Challenging Light Conditions}},
   volume = {22},
   url = {https://www.mdpi.com/1424-8220/22/3/835},
   year = {2022},
}

@article{Wang2022,
  title={{Automated camera-exposure control for robust localization in varying illumination environments}},
  author={Wang, Yu and Chen, Haoyao and Zhang, Shiwu and Lu, Wencan},
  journal={Autonomous Robots},
  volume={46},
  number={4},
  pages={515--534},
  year={2022},
  publisher={Springer}
}

@article{Han2023,
  doi = {10.1109/tmech.2023.3234316},
  url = {https://doi.org/10.1109/tmech.2023.3234316},
  year = {2023},
  publisher = {Institute of Electrical and Electronics Engineers ({IEEE})},
  volume = {28},
  number = {4},
  pages = {2225--2235},
  author = {Bin Han and Yicheng Lin and Yan Dong and Hao Wang and Tao Zhang and Chengyuan Liang},
  title = {{Camera Attributes Control for Visual Odometry With Motion Blur Awareness}},
  journal = {{IEEE}/{ASME} Transactions on Mechatronics (TMECH)}
}

@article{shannon1948mathematical,
  title={A mathematical theory of communication},
  author={Shannon, Claude Elwood},
  journal={The Bell system technical journal},
  volume={27},
  number={3},
  pages={379--423},
  year={1948},
  publisher={Nokia Bell Labs}
}

@inproceedings{KITTI2012,
   author = {A. Geiger and P. Lenz and R. Urtasun},
   booktitle = {IEEE Conference on Computer Vision and Pattern Recognition},
   month = {6},
   pages = {3354-3361},
   title = {{Are we ready for autonomous driving? The KITTI vision benchmark suite}},
   year = {2012},
}

@inproceedings{NorthCampus,
   author = {Nicholas Carlevaris-Bianco and Ryan M. Eustice},
   doi = {10.1109/IROS.2014.6942941},
   isbn = {978-1-4799-6934-0},
   issn = {21530866},
   booktitle = {IEEE/RSJ International Conference on Intelligent Robots and Systems},
   pages = {2769-2776},
   title = {{Learning visual feature descriptors for dynamic lighting conditions}},
   year = {2014},
}

@inproceedings{TartanDrive2022,
   author = {Samuel Triest and Matthew Sivaprakasam and Sean J. Wang and Wenshan Wang and Aaron M. Johnson and Sebastian Scherer},
   doi = {10.1109/ICRA46639.2022.9811648},
   isbn = {978-1-7281-9681-7},
   issn = {10504729},
   booktitle = {IEEE International Conference on Robotics and Automation (ICRA)},
   pages = {2546-2552},
   title = {{TartanDrive: A Large-Scale Dataset for Learning Off-Road Dynamics Models}},
   url = {https://ieeexplore.ieee.org/document/9811648/},
   year = {2022},
}

@inproceedings{Oxford2020,
   author = {Dan Barnes and Matthew Gadd and Paul Murcutt and Paul Newman and Ingmar Posner},
   doi = {10.1109/ICRA40945.2020.9196884},
   isbn = {978-1-7281-7395-5},
   booktitle = {IEEE International Conference on Robotics and Automation (ICRA)},
   month = {5},
   pages = {6433-6438},
   title = {{The Oxford Radar RobotCar Dataset: A Radar Extension to the Oxford RobotCar Dataset}},
   url = {https://ieeexplore.ieee.org/document/9196884/},
   year = {2020},
}

@article{UMA-VI,
   author = {David Zuñiga-Noël and Alberto Jaenal and Ruben Gomez-Ojeda and Javier Gonzalez-Jimenez},
   doi = {10.1177/0278364920938439},
   issn = {0278-3649},
   issue = {9},
   journal = {The International Journal of Robotics Research},
   month = {8},
   pages = {1052-1060},
   publisher = {SAGE Publications Inc.},
   title = {{The UMA-VI dataset: Visual–inertial odometry in low-textured and dynamic illumination environments}},
   volume = {39},
   url = {http://journals.sagepub.com/doi/10.1177/0278364920938439},
   year = {2020},
}

@article{FinnForest2020,
   author = {Ihtisham Ali and Ahmed Durmush and Olli Suominen and Jari Yli-Hietanen and Sari Peltonen and Jussi Collin and Atanas Gotchev},
   doi = {10.1016/j.robot.2020.103610},
   issn = {09218890},
   journal = {Robotics and Autonomous Systems},
   month = {10},
   pages = {103610},
   publisher = {Elsevier B.V.},
   title = {{FinnForest dataset: A forest landscape for visual SLAM}},
   volume = {132},
   url = {https://linkinghub.elsevier.com/retrieve/pii/S0921889020304504},
   year = {2020},
}

@inproceedings{zhang2016benefit,
  title={{Benefit of large field-of-view cameras for visual odometry}},
  author={Zhang, Zichao and Rebecq, Henri and Forster, Christian and Scaramuzza, Davide},
  booktitle={IEEE International Conference on Robotics and Automation (ICRA)},
  pages={801--808},
  year={2016},
}

@inproceedings{engel2016monodataset,
 author = {J. Engel and V. Usenko and D. Cremers},
 title = {A Photometrically Calibrated Benchmark For Monocular Visual Odometry},
 booktitle = {arXiv:1607.02555},
 arxiv = { arXiv:1607.02555},
 year = {2016},
 month = {July},
}

@article{Bergmann2018,
   author = {Paul Bergmann and Rui Wang and Daniel Cremers},
   doi = {10.1109/LRA.2017.2777002},
   issn = {2377-3766},
   issue = {2},
   journal = {IEEE Robotics and Automation Letters},
   month = {4},
   pages = {627-634},
   publisher = {Institute of Electrical and Electronics Engineers Inc.},
   title = {Online Photometric Calibration of Auto Exposure Video for Realtime Visual Odometry and SLAM},
   volume = {3},
   url = {http://ieeexplore.ieee.org/document/8119575/},
   year = {2018},
}

@INPROCEEDINGS{Kubelka2022,
  author={Kubelka, Vladimír and Vaidis, Maxime and Pomerleau, François},
  booktitle={IEEE/RSJ International Conference on Intelligent Robots and Systems (IROS)}, 
  title={Gravity-constrained point cloud registration}, 
  year={2022},
  pages={4873-4879}
}

@INPROCEEDINGS{Jiajun2020,
  author={{Lv, Jiajun and Xu, Jinhong and Hu, Kewei and Liu, Yong and Zuo, Xingxing}},
  booktitle={2020 IEEE/RSJ International Conference on Intelligent Robots and Systems (IROS)}, 
  title={{Targetless Calibration of LiDAR-IMU System Based on Continuous-time Batch Estimation}}, 
  year={2020},
  volume={},
  number={},
  pages={9968-9975},
  doi={10.1109/IROS45743.2020.9341405}}

@inproceedings{lowe1999object,
  title={{Object recognition from local scale-invariant features}},
  author={Lowe, David G},
  booktitle={IEEE International Conference on Computer Vision},
  pages={1150--1157},
  year={1999}
}

@article{opencv_library,
    author = {Bradski, G.},
    journal = {Dr. Dobb's Journal of Software Tools},
    title = {{The OpenCV Library}},
    year = {2000}
}

@inproceedings{tareen2018comparative,
  title={A comparative analysis of sift, surf, kaze, akaze, orb, and brisk},
  author={Tareen, Shaharyar Ahmed Khan and Saleem, Zahra},
  booktitle={IEEE International conference on computing, mathematics and engineering technologies (iCoMET)},
  pages={1--10},
  year={2018}
}

@inproceedings{sturm2012benchmark,
  title={A benchmark for the evaluation of RGB-D SLAM systems},
  author={Sturm, J{\"u}rgen and Engelhard, Nikolas and Endres, Felix and Burgard, Wolfram and Cremers, Daniel},
  booktitle={IEEE/RSJ international conference on intelligent robots and systems},
  pages={573--580},
  year={2012}
}

@article{nister2004efficient,
  title={An efficient solution to the five-point relative pose problem},
  author={Nist{\'e}r, David},
  journal={IEEE transactions on pattern analysis and machine intelligence},
  volume={26},
  number={6},
  pages={756--770},
  year={2004},
  publisher={IEEE}
}

@article{Baril2022,
  doi = {10.55417/fr.2022050},
  url = {https://doi.org/10.55417/fr.2022050},
  year = {2022},
  month = mar,
  publisher = {Field Robotics Publication Society},
  volume = {2},
  number = {1},
  pages = {1628--1660},
  author = {Dominic Baril and Simon-Pierre Desch{\^{e}}nes and Olivier Gamache and Maxime Vaidis and Damien LaRocque and Johann Laconte and Vladim{\'{\i}}r Kubelka and Philippe Gigu{\`{e}}re and Fran{\c{c}}ois Pomerleau},
  title = {Kilometer-scale autonomous navigation in subarctic forests: challenges and lessons learned},
  journal = {Field Robotics}
}

@inproceedings{detone2018superpoint,
  title={Superpoint: Self-supervised interest point detection and description},
  author={DeTone, Daniel and Malisiewicz, Tomasz and Rabinovich, Andrew},
  booktitle={Proceedings of the IEEE conference on computer vision and pattern recognition workshops},
  pages={224--236},
  year={2018}
}
